\documentclass{article}

\usepackage{PRIMEarxiv}

\usepackage[utf8]{inputenc} 
\usepackage[T1]{fontenc}    
\usepackage{hyperref}       
\usepackage{url}            
\usepackage{booktabs}       
\usepackage{amsfonts}       
\usepackage{nicefrac}       
\usepackage{microtype}      
\usepackage{lipsum}
\usepackage{fancyhdr}       
\usepackage{graphicx}     
\usepackage{algorithm}
\usepackage{algpseudocode}
\usepackage{amsmath}
\usepackage{placeins}
\graphicspath{{media/}}

\title{Autoencoder based approach for the mitigation of spurious correlations
}

\author{
  Srinitish Srinivasan \\
  Vellore Institute of Technology \\
  India\\
   \And
Karthik Seemakurthy\\
  University of Lincoln \\
  United Kingdom\\
}

\begin{document}
\maketitle

\begin{abstract}
Deep neural networks (DNNs) have exhibited remarkable performance across various tasks, yet their susceptibility to spurious correlations poses a significant challenge for out-of-distribution (OOD) generalization. Spurious correlations refer to erroneous associations in data that do not reflect true underlying relationships but are instead artifacts of dataset characteristics or biases. These correlations can lead DNNs to learn patterns that are not robust across diverse datasets or real-world scenarios, hampering their ability to generalize beyond training data. In this paper, we propose an autoencoder-based approach to analyze the nature of spurious correlations that exist in the Global Wheat Head Detection (GWHD) 2021 dataset. We then use inpainting followed by Weighted Boxes Fusion (WBF) to achieve a 2\% increase in the Average Domain Accuracy (ADA) over the YOLOv5 baseline and consistently show that our approach has the ability to suppress some of the spurious correlations in the GWHD 2021 dataset. The key advantage of our approach is that it is more suitable in scenarios where there is limited scope to adapt or fine-tune the trained model in unseen test environments.
\end{abstract}


\section{Introduction}
Object detection, a fundamental task in computer vision, plays a pivotal role in various real-world applications such as autonomous driving, surveillance, and medical imaging~\cite{zou2023object}. Deep learning-based object detection methods, particularly those employing convolutional neural networks (CNNs), have achieved remarkable success in accurately localizing and classifying objects within images~\cite{liu2020deep}. However, despite their proficiency, these models are susceptible to a critical challenge: spurious correlations~\cite{wang2024learning}.

Spurious correlations refer to misleading associations between input features and target labels that are not indicative of the true underlying relationships. These correlations often arise due to confounding factors, dataset biases, or the complex nature of real-world data~\cite{ye2024spurious}. In the context of object detection, spurious correlations can manifest in various forms, leading to erroneous predictions, degraded model performance, and compromised generalization capabilities. Training datasets may inadvertently contain biases in terms of object distribution, background clutter, or imaging conditions, leading the model to learn and exploit spurious correlations rather than true object features. Moreover, the complex interplay between object context and appearance can introduce spurious correlations that are difficult for models to disentangle and thus leading to false positive detection's. In this paper, we propose an approach to suppress the spurious correlations in agricultural domain. 

The wide spread adoption of deep learning usage for precision agriculture~\cite{nasir2021deep} ~\cite{ferentinos2018deep}, increased the need for effective mitigation of spurious correlations during inference time. Incorrect predictions such as false positives or false negatives on sensitive problems such as crop-disease detection and crop-weed segmentation can prove to be extremely costly. This may not only negatively impact the farmers and industries financially but also the health and well-being of consumers. With respect to wheat heads, precise deep learning techniques that can detect all types of wheat heads will considerably be able to reduce the cost of production and increase overall crop yield thus raising profits. Such detections may also correlate to early 

\begin{figure}
    \centering
    \begin{tabular}{cc}
   \includegraphics[scale=0.15]{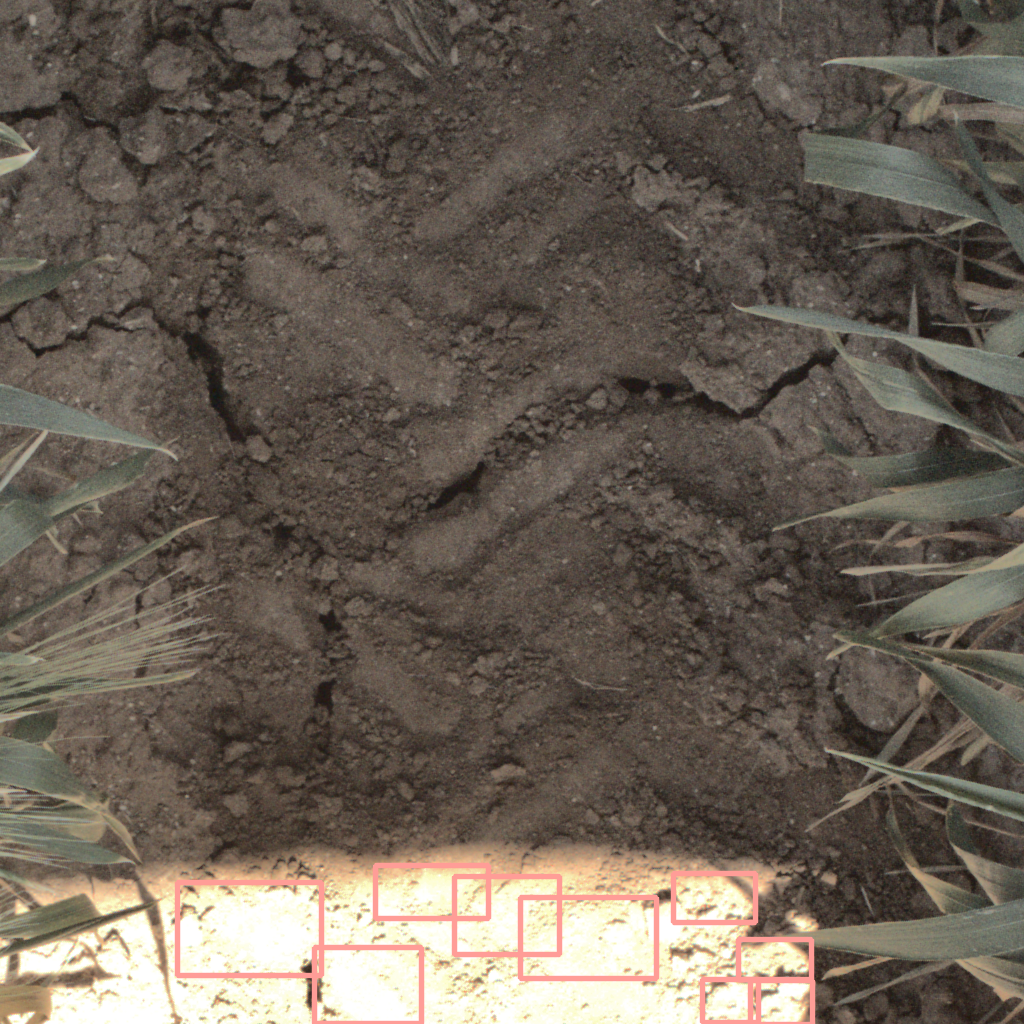} &
   \includegraphics[scale=0.15]{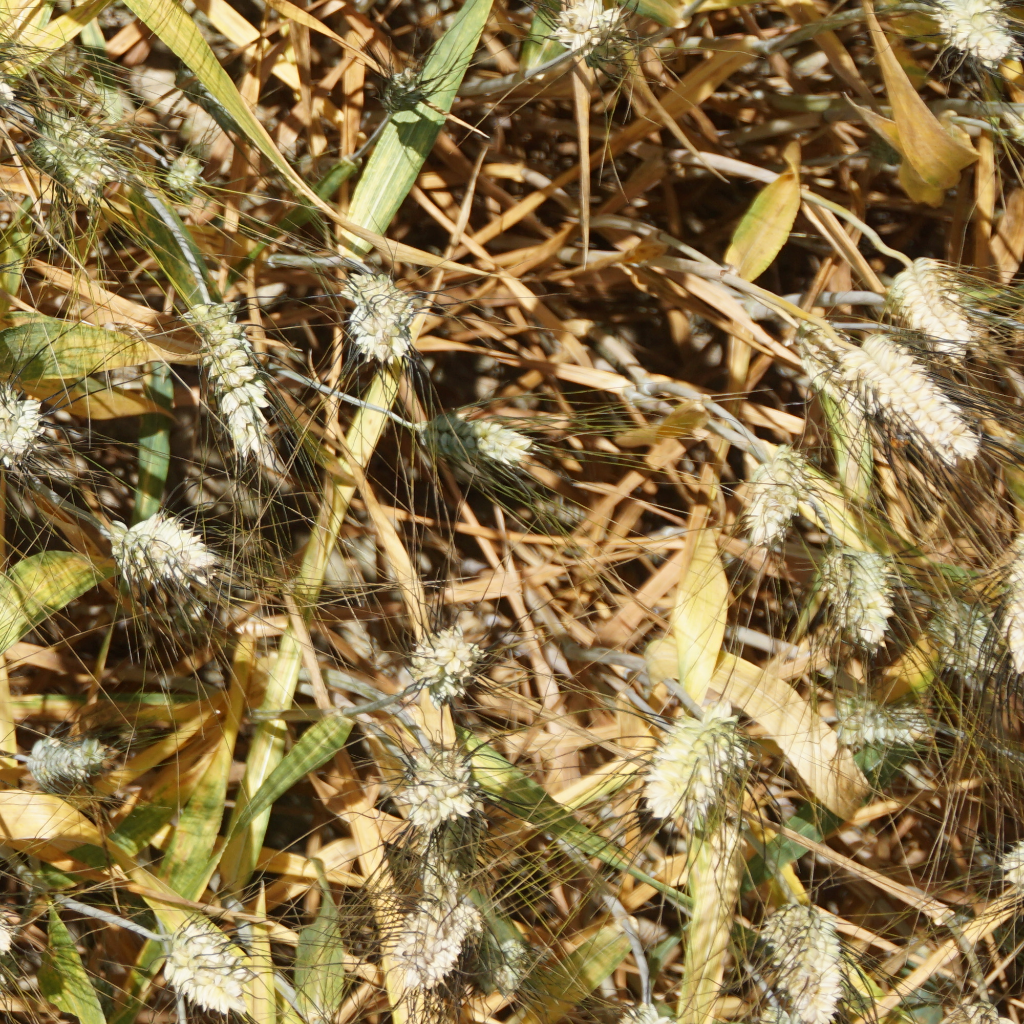} \\  
   (a) & (b) \\
   \end{tabular}
   \label{fig:demo_spurious_correlations}
   \caption{Sample spurious correlations on GWHD 2021 dataset. (a) YoloV5 detections on the test sample, and (b) Sample image from training set.}
\end{figure}

In this work, we will demonstrate the ability to identify and suppress the spurious correlations in the Global Wheat Head Detection (GWHD) 2021 dataset~\cite{david2023global} using YoloV5 detector. Fig.~\ref{fig:demo_spurious_correlations} (a) shows the sample false detections that we address in this paper. It can be seen that the bright sunny regions are identified as false wheat heads. One of the key reasons behind these false detections are due to the nature of wheat heads in the training set. Fig.~\ref{fig:demo_spurious_correlations} (b) shows the sample image from GWHD 2021 traning set and it can be clearly seen that the visual signatures of wheat heads are very similar to that of the bright sunny regions and this is the kind of spurious correlations that we address in this work. Although our approach is detector agnostic, the reason behind the choice of YoloV5 detector is because of its wide usage by majority of the winning solutions in GWHD 2021 competition~\cite{david2023global}.

Unlike the existing techniques which mitigate spurious correlations during the training/fine-tuning phase~\cite{yang2023mitigating}, our approach is applied during inference-time in three stages. In the first stage, we initially identify bright sunny regions in GWHD 2021 dataset through thresholding and inpaint those regions using Navier-Stokes approach. We then use an autoencoder based architecture to minimise the effects of spurious correlations. In the final stage, the original detections are fused along the detections obtained in the second stage using Weighted Boxes Fusion (WBF). By using qualitative and quantitative experiments, we demonstrate the effectiveness of our proposed approaches in uncovering and mitigating spurious correlations, thereby advancing the state-of-the-art wheat detection and paving the way towards more reliable and interpretable computer vision systems.

The key contributions of the proposed approach are as follows:
\begin{itemize}
    \item We propose an explainable approach to localise the regions of spurious correlations for GWHD 2021 dataset. 
    \item We initally use the inpainting followed by an autoencoder architecture to minimise the effects of these artefacts.  
    \item We finally propose to use Weighted Box Fusion(WBF) to effectively suppress these spurious correlations.  
    \item Our proposed strategy results in 2\% improvement in Average Domain Accuracy(ADA) on the test split of GWHD 2021 dataset. 
\end{itemize}

The sections of the paper is divided as follows: Section \ref{sec:relatedwork} discusses previous works on enhanced YOLO architecture for wheat head detection and spurious correlations in machine learning models which are closely related to our study. Section \ref{sec:proposed_approach} discusses the detailed steps involved in the proposed approach followed by the results and discussion in Section \ref{sec:results} where we present both the quantitative and qualitative analysis of our approach. We finally conclude our paper in Section~\ref{sec:conclusion}. 

\section{Related Work}\label{sec:relatedwork}
\subsection{Wheat Head Detection}
Several researchers have used YOLO models in their enhanced form tailored for the wheat head domain~\cite{david2023global}.~\cite{shen2023yolov5} uses an enhanced form of YOLOv5 by changing the backbone architecture of YOLOv5 to improve detections on wheat head images. The researchers reduce the number of Cross Stage Partial (CSP) modules, they replace the convolutions of the CSP module with separable convolutions which is further added to the fusion path in order to reduce the redundant information in the feature map. Further, they also add the co-attention mechanism and report state-of-the-art results on this approach.~\cite{han2022wheat} used YOLOv5 with Weighted Coordinate Attention (WCA) to aggregate features along x and y coordinates thus capturing long range dependencies.~\cite{zhang2022high} designed specific loss functions to focus more on the wheat head spikes of the images rather than the background. The researchers replaced the LeakyRELU activation function of the backbone with the Mish activation function. Further, they also added a label smoothing function at the output of the backbone network to prevent classification overfitting. ~\cite{meng2023yolov7} proposed the YOLOv7-MA model where they introduce micro-scale detection layers and the convolution block attention module which enhances the information mapping to the wheat head spike region and weakens background information. 

\subsection{Spurious Correlations}
\label{sec:spurious_correlations}
The presence of spurious correlations is one of the biggest challenges in machine learning models. Domain generalization, invariant learning, group robustness, shortcut learning, and simplicity bias are several other alternative names for spurious correlations~\cite{ye2024spurious}. According to the survey in~\cite{ye2024spurious}, all the approaches which attempt to mitigate spurious correlations can be broadly categorised into three different classes. Data manipulation based approaches modify the input data to improve the sample diversity which will aid in reducing the impact of spurious correlations~\cite{srivastava2020robustness}\cite{puli2022nuisances}\cite{yao2022improving}. Another class of techniques which deal with these artefacts comes under the category of representation learning and these approaches deal with improving model robustness~\cite{wang2021causal}~\cite{agarwal2020towards}~\cite{sun2024beyond}. Invariant learning are the class of approaches which deal with penalising the models and assist them to avoid overfitting~\cite{arjovsky2019invariant}~\cite{krueger2021out}.        

Most of the aforementioned approaches, which operate on images, are primarily focused on suppressing spurious correlations in the context of image classification. In this paper, we present our approach to suppress spurious correlations for object detection application. The key contribution of this work is that we initially localize the regions where possible spurious correlations exist in the form of masks and then utilize these masks to suppress the artefacts through inpainting and WBF. 

\section{Proposed approach} \label{sec:proposed_approach}
In this section, we describe the detailed steps of our proposed approach. We initially describe the nature of spurious correlations that we encounter in GWHD 2021 dataset followed by an inpainting procedure to suppress these artefacts. We then describe the details of the proposed autoencoder architecture for YoloV5 backbone which will be used to smoothen the image followed by the details of Weighted Boxes Fusion (WBF) to obtain final enhanced wheat head predictions.   

\subsection{Preliminaries}\label{sec:patterns}
\begin{figure}[!h]
        \centering
        \caption{False positives in the test split of GWHD 2021 dataset.}

        \begin{tabular}{ccc}
             \includegraphics[width=.33\linewidth]{Incorrect_predictions/figure_1.png} &
             \includegraphics[width=.33\linewidth]{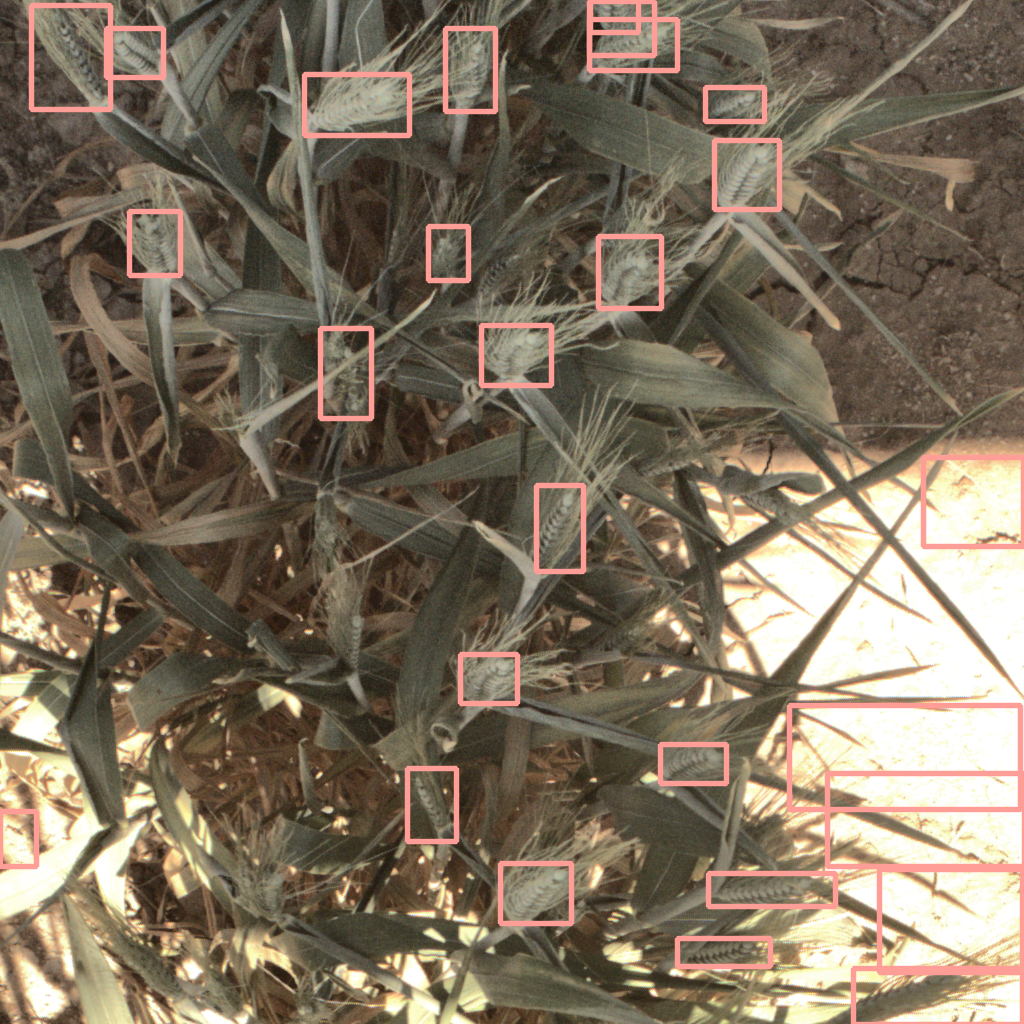} &
             \includegraphics[width=.33\linewidth]{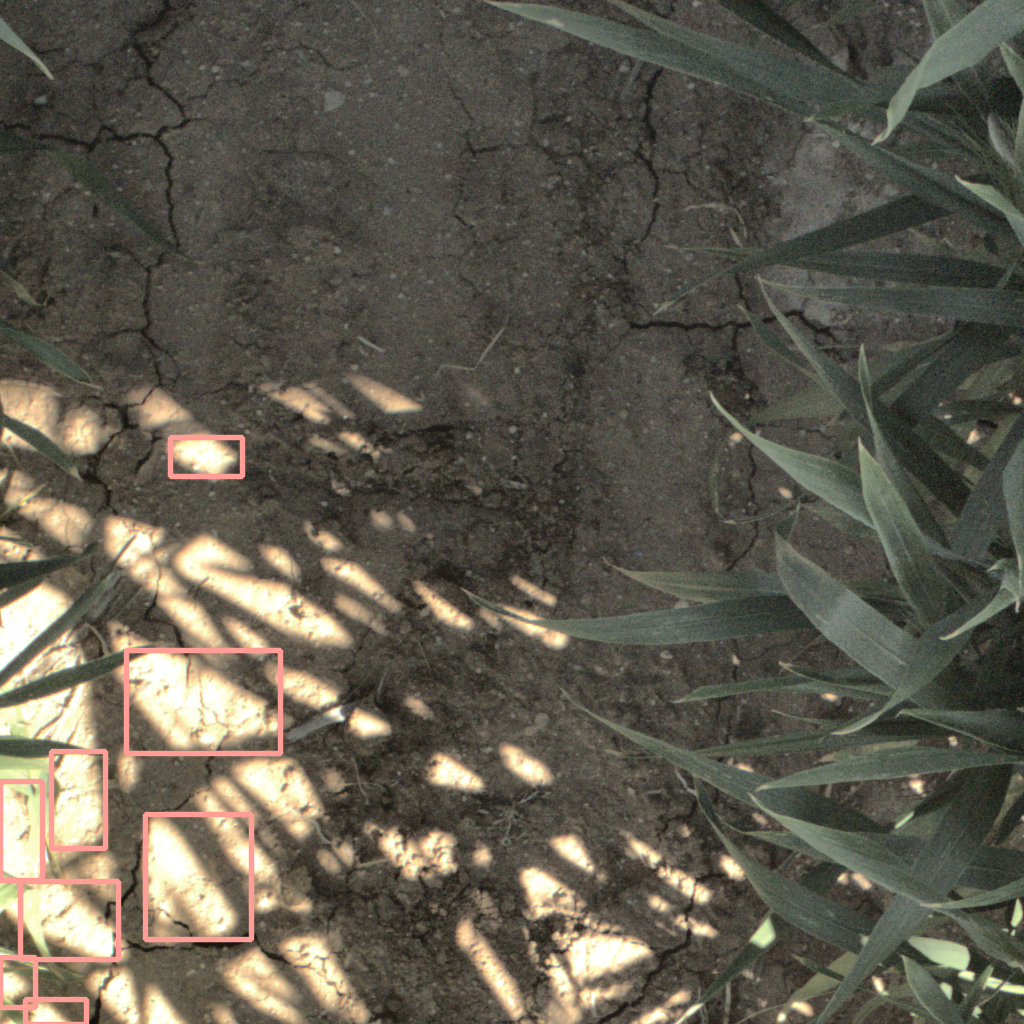}
        \end{tabular}
        \label{fig:mis-outputs}
\end{figure}

YoloV5 has been the popular choice among the winning solutions of GWHD 2021 competition~\cite{david2023global} where most of the approaches focused on improving the overall average domain accuracy (ADA) on the test split of the dataset. None of these approaches dealt with artefacts specific to the GWHD 2021 dataset. Similar to spurious correlations reported in Fig.~\ref{fig:demo_spurious_correlations}, we observed a few artefacts specific to GWHD 2021 dataset. On images trained on a darker background, the model was able to detect the wheat heads with a high accuracy. Further, wheat heads that were light-green to yellow i.e relatively brighter than the background of the image were detected to a high precision as well. However, the images from test split which had high glare due to sunlight forming shadows were detected as false wheat heads. Such images with high sunlight regions forming shadows or heavy glare had several bounding box predictions on regions with no wheat heads, grassland or vegetation. Fig.~\ref{fig:mis-outputs} represents a set of few instances on the test set where false positives have occurred on bright sunlight regions cast on the soil which were typically observed only on the test set.  
In this paper, our aim is to suppress these artefacts by conducting several experiments which not only suppress these miss-classifications during inference time but also preserve the true-positive predictions. 

\subsection{Explanation of Misclassified Instances using Autoencoders based on YOLOv5 Backbone}\label{sec:autoencoder}

\begin{figure}[!h]
        \caption{Architectural diagram of autoencoder based architecture to explain test set predictions.}
         \includegraphics[width=\linewidth,scale=2]{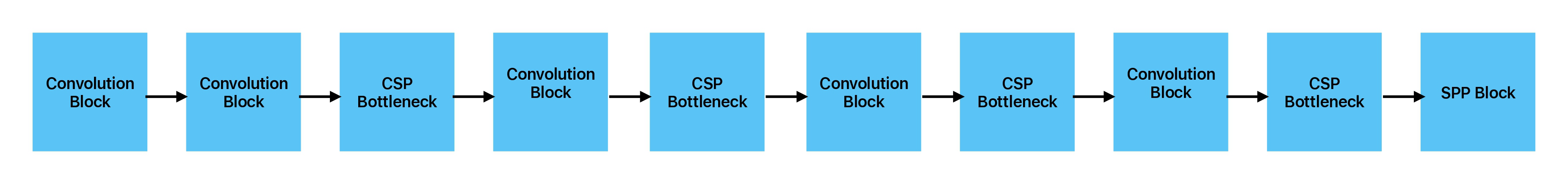}

        \label{fig:backbone}
\end{figure}

In this section, we provide the details of autoencoder architecture for YoloV5 backbone that we used to explain the misclassified instances.   
It is to note that the autoencoder was trained on the same train set as the YOLOv5s. Fig.~\ref{fig:backbone} presents the proposed autoencoder architecture, where CSP Block represents the Cross Stage Partial Convolutions as used in YOLOv5 backbone and SPP refers to Spatial Pyramid Pooling which is employed in the final layer of the backbone. The decoder constructed for this task is symmetrical to the encoder backbone which was made possible with symmetrical transpose convolutions and up-sampling operations. Further, we also include skip connections between the encoder and decoder to prevent vanishing gradients and retain underlying information from earlier layers. We propose a mask based loss function specifically for our task to focus only on the object of interest and is given in Eq.~\ref{eq:auto_loss}

\begin{equation}
    \label{eq:auto_loss}
    L=\Sigma_{i=1}^{i=N}\text{MSE}(y_i,\hat y_i)\odot P
\end{equation}
where $P$ represents the penalty matrix which has the same dimensions as that of the image, MSE refers to Mean Square Error and $\odot$ denotes the element-wise multiplication. Regions belonging to a bounding box are given more weight than the other regions. We use inverse weights to weigh pixels that belong to a bounding box. Eq.~\ref{eq:weight} describes the creation of the penalty matrix $P$. This ensures that the model focuses more on the wheat head spikes than the background during reconstruction. 

\begin{align}
    \label{eq:weight}
    P_i=\begin{cases}
          \frac{1}{0.01}&\text{if }m_i=1 \\
            \frac{1}{0.99}& \text{otherwise}
        \end{cases}
\end{align}

where $m_i$ is a binary variable that equates to 1 if the corresponding pixel belongs to a bounding box, else it is 0. The outputs of the reconstructed images explain the patterns of the misclassified instances. Apart from wheat heads, which is the object of interest, the regions of high sunlight have been activated as well. This indicates that the transfer of information from train to test set includes only the shape and spectral features but not some of the underlying patterns that may represent the wheat head this providing a model-based explanation for the misclassified instances.

\begin{figure}
    \centering
    \caption{Reconstructed Images of pixels activated for predicted wheat head spike regions in Fig.~\ref{fig:mis-outputs}}
    
    \begin{tabular}{ccc}
         \includegraphics[width=.33\linewidth]{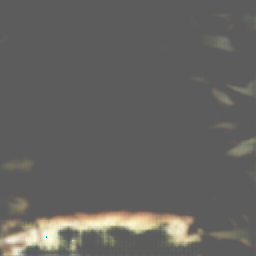}&
        \includegraphics[width=.33\linewidth]{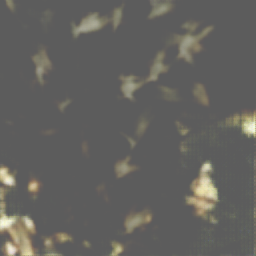}&
        \includegraphics[width=.33\linewidth]{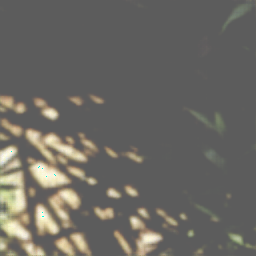} \\
    \end{tabular}
    \label{fig:auto_output}
\end{figure}

\subsection{Image Inpainting}\label{sec:inpaint}
The first step of our proposed approach is to sparsely localise the presence of spurious correlations. As mentioned in the previous section, the bright sunny regions are responsible for performance degradation. By using inpainting on the bright sunlight regions, we suppress the impact of saturated regions on the model performance. Algorithm~\ref{alg:inpaint} summarises the workflow of the in-painting procedure. Among the various techniques that we investigated~\cite{bertalmio2001navier}\cite{telea2004image}, Navier Stokes in-painting algorithm was shown to produce better results for our scenario. We consider high brightness regions with pixel values ranging from 170 to 255 as the threshold region for constructing a binary mask. We use 9 $\times$ 9 kernels for Gaussian Blur followed by image erosion and dilation for 2 and 4 iterations, respectively. Based on the mask obtained, image inpainting is performed using the  Navier Stokes algorithm. In algorithm \ref{alg:inpaint}, $P$ refers to the image path, $I$ is the image matrix and $M$ refers to the binary mask. 

\begin{algorithm}
    \caption{Inpainting Procedure}\label{alg:inpaint}
    \begin{algorithmic}
        \State $I \gets Read(P)$;
        \State $M \gets Gray(I)$;
        \State $M \gets Blur(M)$;

        \State $N \gets 1$
        \While{$N \leq 6$}
            \If{$N$<=2}
                \State $M \gets Erode(M)$;
            \ElsIf{$N$>2}
                \State $M \gets Dilate(M)$;
        \EndIf
        \EndWhile
        
        \State $I \gets Mask(M)$;
        \State $I \gets Inpaint(I,M)$;
    \end{algorithmic}
\end{algorithm}

\subsection{Glare removal using autoencoder}\label{sec:auto}
\begin{figure}[!h]

        \caption{Architecture Diagram of autoencoder for image glare correction.}
         \includegraphics[width=\linewidth,scale=2]{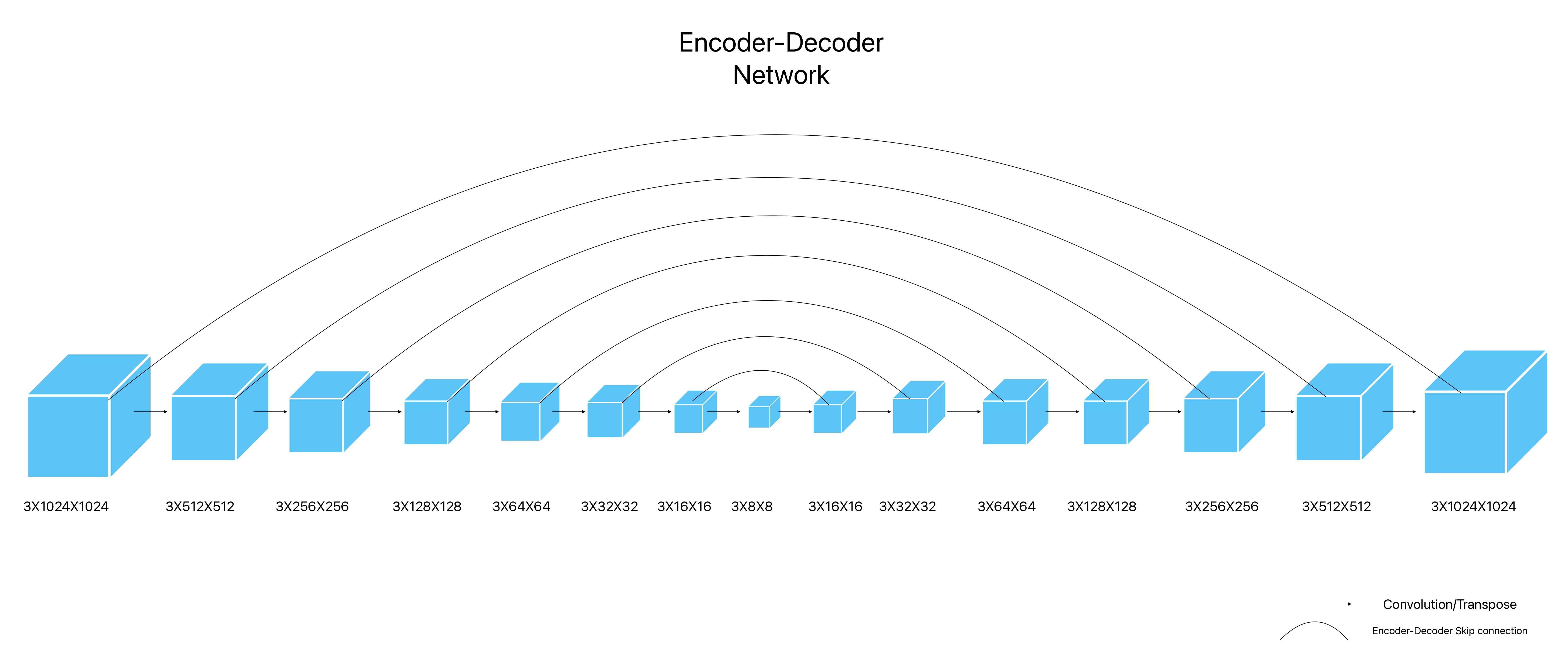}

        \label{fig:autoencoder}
\end{figure}

Although inpainting process described in Sec.~\ref{sec:inpaint} suppresses the artefacts, they tend to introduce minor discontinuities wherever there is significant intensity changes. In order to smoothen these discontinuities, we train SegNet~\cite{badrinarayanan2017segnet} inspired autoencoder that reconstructs images with comparatively lesser brightness to prevent miss-classification based on the patterns mentioned in the earlier sections. Fig.~\ref{fig:autoencoder} represents the architecture used for this experiment. We make use of 9 double-strided convolution layers with 3X3 kernels. The reconstruction block uses transpose convolutions of same kernel sizes. Further skip connections were employed between encoder and decoder blocks to prevent vanishing gradients and account for long-term dependencies. The results of the Navier Stokes procedure described in algorithm \ref{alg:inpaint} forms the target reconstructed outputs of the autoencoder while the input remains constant. Eq.~\ref{eq:autoencoder} describes the functional task of the autoencoder. 
\begin{equation}
        \label{eq:autoencoder}
            M(X)=Y
\end{equation}
where $M$ refers to the autoencoder, $X$ is the input image with high glare and shadow forming regions resembling the shape of wheat heads and $Y$ is the target reconstructed output formed as a result of algorithm \ref{alg:inpaint}. Eq.~\ref{eq:loss} describes the objective function used to train the autoencoder architecture.
\begin{align}
        \label{eq:loss}
         smooth_{L_1}(x)=\begin{cases}
                            \frac{0.52}{\beta}&\text{if }|x|<\beta\\
                            |x|-0.5*\beta & \text{otherwise}
                        \end{cases}
\end{align}
where $x$ refers to pixel-wise difference between the predicted and target values. 

\subsection{Weighted Boxes Fusion}\label{sec:WBF}
Finally, we perform Weighted Box Fusion(WBF)~\cite{weighted} on the predictions of each method discussed in Sec.~\ref{sec:inpaint} and Sec.~\ref{sec:auto}. Algorithm \ref{alg:bbx} represents the final workflow to improve the bounding box predictions on the wheat heads. Each set of instances obtained after In-painting and Autoencoder Reconstruction were passed into the previously trained YOLOv5 model for inference. The bounding box predictions obtained after this stage is fused with the original set of predictions. Further, we also perform fusion experiments by including both domain specific weights of YOLOv5 for this task along with the weights obtained on fine-tuning the ImageNet Weights as described in Sec.~\ref{sec:quant}.

\begin{algorithm}
    \caption{Proposed Workflow}\label{alg:bbx}
    \begin{algorithmic}

        \State $I \gets Read(P)$;
        \State $I^{'} \gets Inpaint(I)$;
        \State $I^{''} \gets Autoencoder(I)$;    \Comment{Get the 3 sets of Instances for Box Fusion}

        \State $B \gets Yolo(I)$;
        \State $B^{'} \gets Yolo(I^{'})$;
        \State $B^{''} \gets Yolo(I^{''})$; \Comment{Perform YOLO Inference on all sets of instances}
        
        \State $F \gets Wbf(B,B',B'')$ \Comment{Output Final Set of Predictions}
        
    \end{algorithmic}
\end{algorithm}

\section{Results and Discussion}\label{sec:results}
\subsection{Hardware Environment}
{
    All experiments were conducted on a 2022 Macbook Pro. The configurations of the M2 Macbook Pro are inclusive of a 8GB unified memory with a 256 GB hard disk storage. The number of cores of the CPU and GPU are 8 and 10, respectively. The average time taken to complete 75 epochs of training of YOLOv5 on the CPU was 10 hours. The Autoencoder was trained on the GPU whose average time of completion was 1 hour. 
}
\subsection{Dataset}
{
    All experiments and evaluations were conducted on the Global Wheat Head Dataset(GWHD), 2021\cite{global}. The dataset is composed of more than 6000 images of resolution 1024X1024 with more than 3000 unique wheat heads and bounding boxes provided for each one of them. The training set comprises of images obtained from Europe and Canada  amounting to 4000 images. The final evaluation dataset for the competition comprises of 2000 images obtained from North America (except Canada), Asia, Africa and Oceania. The several new measurement techniques and the wide range of regions the dataset represents makes it a multi-distributional dataset which is ideal for our study. 
}

\subsection{Training Environment of Models used}
    \subsubsection{YOLOv5 Training Environment}\label{sec:yolov5}
    {
    We initially train a YOLOv5s from scratch on the Global Wheat Dataset, 2021. The model was trained for 100 epochs with a batch size of 8. Adam optimizer~\cite{kingma2014adam} was used for optimization with a cyclic learning rate policy. The Binary cross entropy function was used to classify the classes of the instances for each image and Mean Square Error(MSE) was used for predicting the bounding box coordinates. 
    }
    
    \subsubsection{Autoencoder parameters for Glare Removal}
    {
     The training parameters used for training the autoencoder architecture for reducing glare as described in Sec.\ref{sec:auto} are as follows: The learning rate was set to 0.002 with a weight decay of 0.001. The batch size was considered to be 8. We trained the model for upto 10,000 epochs using Adams~\cite{kingma2014adam} optimizer with Beta 1 and Beta 2 as 0.9 and 0.999 respectively. 
     }
    
    \subsubsection{Autoencoder Environment for Test Set Detection Explanations}
    {
        The Autoencoder based on YOLOv5 Backbone as discussed in Sec. ~\ref{sec:autoencoder} was trained for 2000 epochs. The initial learning rate was set to 0.001 which was altered during training based on the Cosine Annealing learning rate policy. The batch size was considered to be 8. Adams~\cite{kingma2014adam} optimizer was used for training with Beta 1 and Beta 2 set to 0.9 and 0.999 respectively. 
    }

\subsection{Quantitative Evaluation}\label{sec:quant}
{
    \paragraph{Evaluation Metrics}
    \paragraph{}
        The metric used for the quantitative comparison of our proposed methodology is the Average Domain Accuracy(ADA). For one image the accuracy is computed as described in Eq.~\ref{accuracy_eqn}

        \begin{equation}
            \label{accuracy_eqn}
             A=\frac{TP}{TP+FP+FN}
        \end{equation}
        
        where, $TP$ refers to True Positive which indicates that the ground truth bounding box has matched with one predicted box, $FP$ refers to False Positive where a predicted box has not matched with any ground truth box and $FN$ refers to False Negative where the ground truth box matches no box.

    \paragraph{Matching Criteria}
    \paragraph{}
        Two Boxes are said to be matched if their Intersection over Union(IOU) is higher than a threshold of 0.5. 

    \paragraph{Average Domain Accuracy}
    \paragraph{}
    The accuracy's of all images in a particular domain is averaged to give the Average Accuracy for that domain. The average accuracy of all domains refers to the Average Domain Accuracy(ADA) and is computed using Eq.~\ref{ADA}.

    \begin{equation}
        \label{ADA}
        \text{ADA}=\frac{\Sigma_{i=1}^{i=N}A_i}{N}
    \end{equation}
    where, $A_i$ refers to the Accuracy of domain $i$ and $N$ is the total number of domains.

    \paragraph{Quantitative Discussion}
    \paragraph{}
    Tables \ref{Ada_1} and \ref{Ada_2} describe the Average Domain Accuracy's obtained for our proposed approach. The experiments were tested for three sets of confidence scores namely 25\%, 30\% and 35\%. The original score on evaluating the YOLOv5 was 0.580, 0.581 and 0.583 for the 3 confidence scores, respectively. On performing experiments with the combination of the outputs of the in-painting and autoencoder mitigated predictions through WBF, we notice a significant improvement in the ADA. From 0.58 on the original YOLOv5, the ADA improved by nearly 1 percent to 0.59.

    Performing WBF with the trained weights of the YOLOv5 model for this task resulted in an additional improvement to the Average Domain Accuracy (ADA). We report a 2\% increase in the ADA by combining the predictions of the outputs of the mitigation techniques and the trained weights. We also perform experiments by taking only combinations of two sets of predictions. The results after the ensemble of image in-painting predictions resulted in a slight improvement of ADA scores. Further, on comparing the results with the ensemble of the instances corrected after passing them through the autoencoder architecture that reconstructs instances after correcting the glare factor, we obtain the same performance increase as the in-painted instances. From a quantitative point of view, the experiments we performed to validate our proposed approach to correct misclassified instances indicate an improvement from the initial results. Table \ref{tab:overall} describes the overall average relative performance of our proposed methodology considering all confidence scores and techniques.    
}

{

\begin{table}
\caption{ADA Scores for Test without Weight Ensemble}
\label{Ada_1}
\centering
\begin{tabular}{|l|l|l|l|l|}
\hline
 Confidence&  
           Original &  
          Original+ Inpaint & 
          Original+ Autoencoder & 
          Original+Inpaint+Autoencoder\\
\hline
          0.30 & 0.581 & 0.584 & 0.582&\textbf{0.588}\\
          0.35 & 0.583 & 0.585 & 0.581 &\textbf{0.587}\\
          0.25 & 0.580 &  0.582 & 0.580 &\textbf{0.588}\\
\hline
\end{tabular}
\end{table}

\begin{table}
\caption{ADA Scores for Test Set with Weight Ensemble}
\label{Ada_2}
\centering
\begin{tabular}{|l|l|l|l|l|}
\hline
 Confidence& 
           Original &  
          Original+ Inpaint & 
          Original+ Autoencoder & 
          Original+Inpaint+Autoencoder\\
\hline
         0.30 & 0.581&\textbf{0.594}&0.592&0.587\\
        0.25 & 0.580 &\textbf{0.599}&0.594&0.592\\
\hline
\end{tabular}
\end{table}

\begin{table}
    \caption{Overall Average Comparative Performance}
    \label{tab:overall}
    \centering
    \begin{tabular}{|l|l|}
    \hline
    Model&ADA Score\\
    \hline
         YOLOv5&0.581  \\
         WBF with In-painting Predictions & 0.594\\
         WBF with Autoencoder Predictions & 0.592\\
         \textbf{WBF with all predictions combined} & \textbf{0.599}\\
    \hline
    \end{tabular}
\end{table}
}

   \begin{figure}
            \centering
            \caption{Qualitative analysis of proposed approach.}
    
            \begin{tabular}{|c|c|c|c|}
                \hline
                \textbf{Model}&\textbf{Sample 1}&\textbf{Sample 2}&\textbf{Sample3 }\\ \hline
            
                \textbf{YOLOv5}&\includegraphics[width=.25\linewidth]{Incorrect_predictions/figure_1.png} &  \includegraphics[width=.25\linewidth]{Incorrect_predictions/figure_2.png} & \includegraphics[width=.25\linewidth]{Incorrect_predictions/figure_3.png} \\ \hline

                \textbf{Autoencoder}&\includegraphics[width=.25\linewidth]{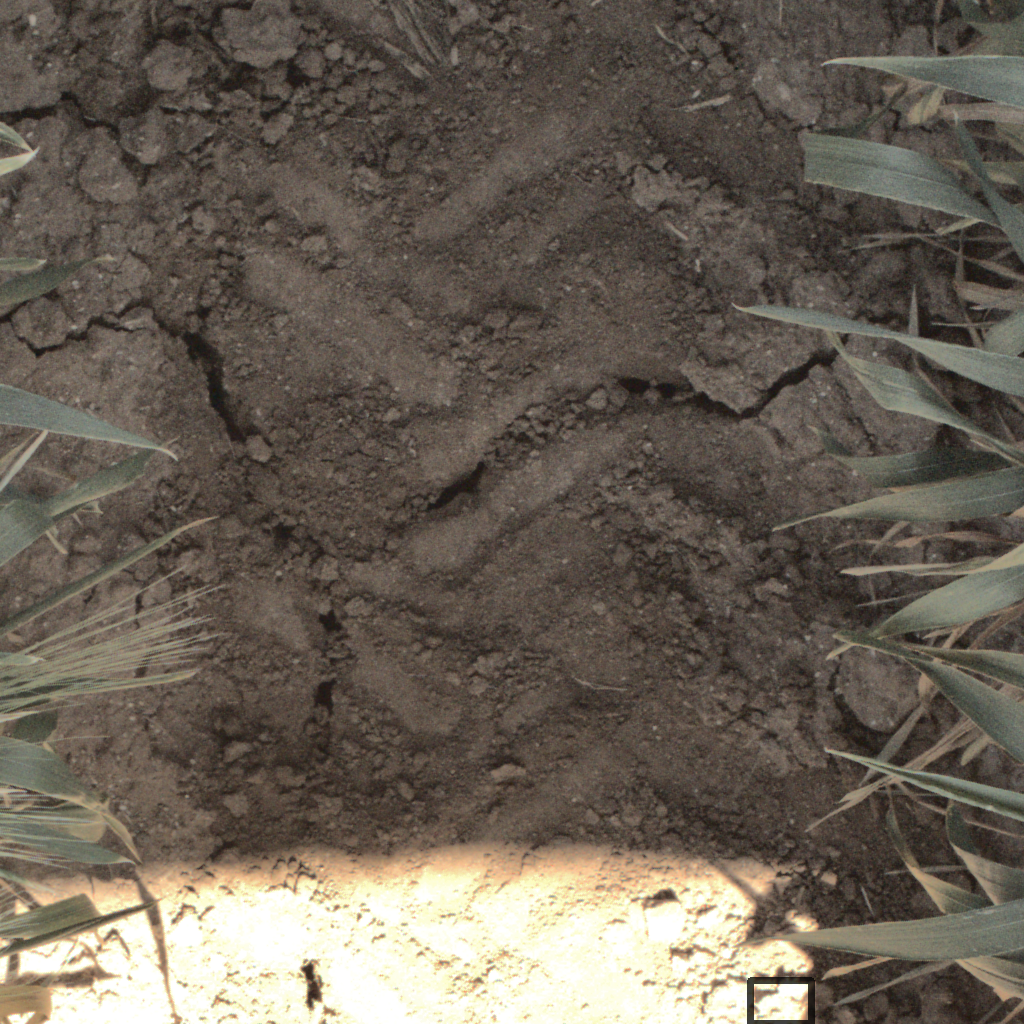} & \includegraphics[width=.25\linewidth]{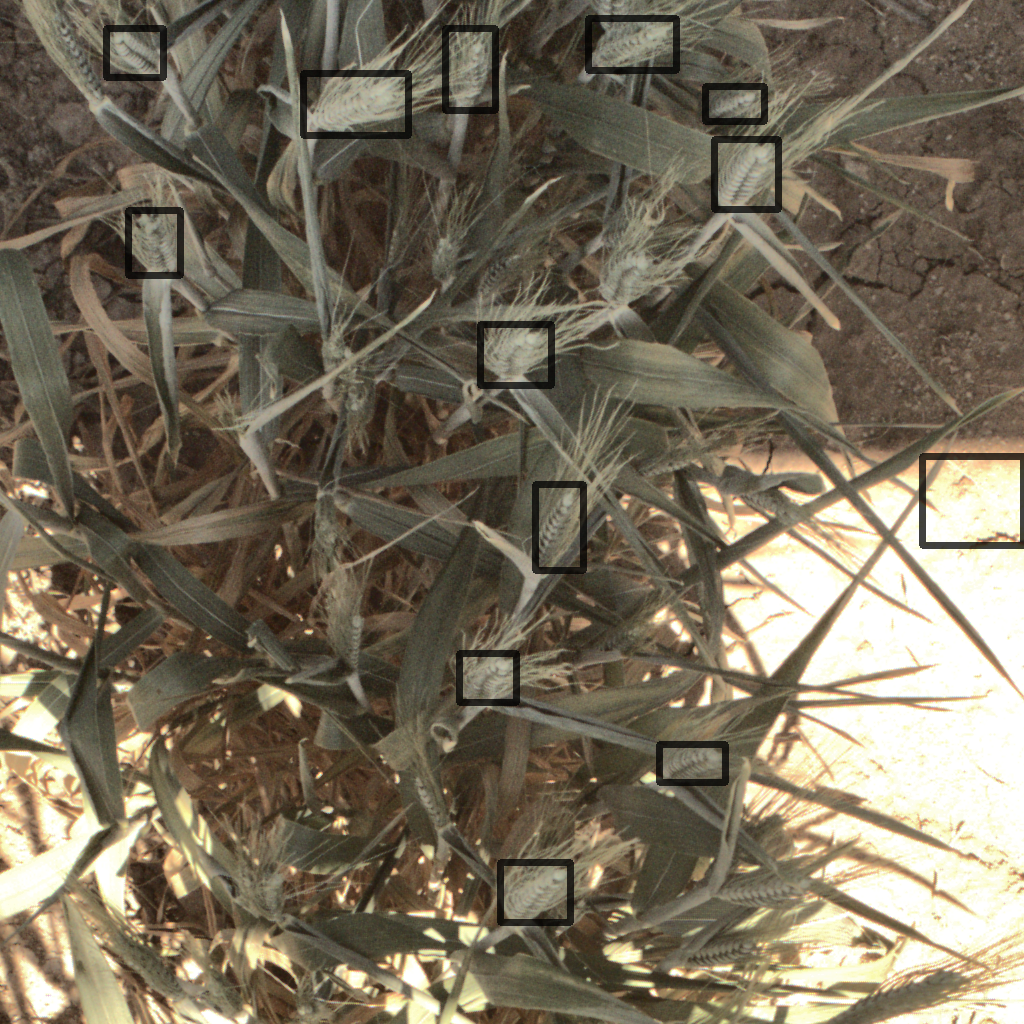} & \includegraphics[width=.25\linewidth]{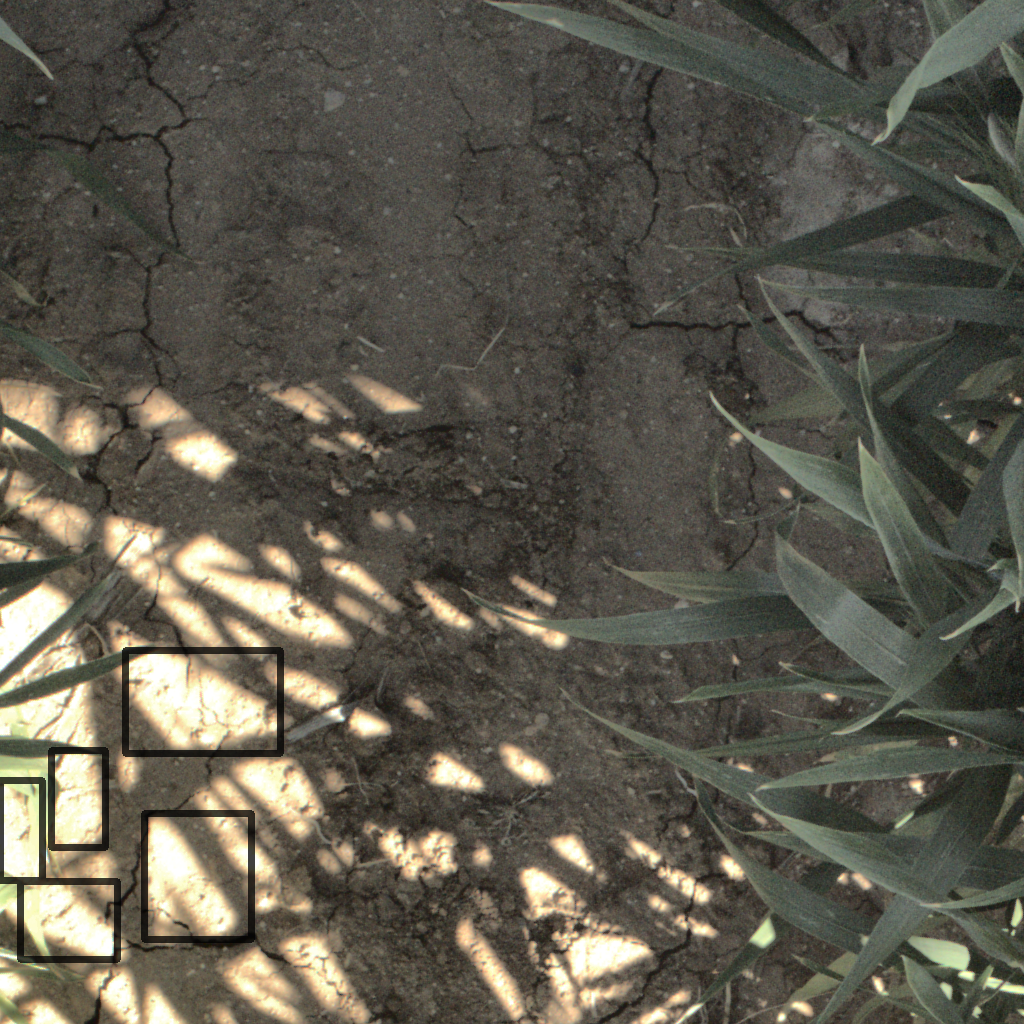} \\ \hline

                \textbf{Inpainting}&\includegraphics[width=.25\linewidth]{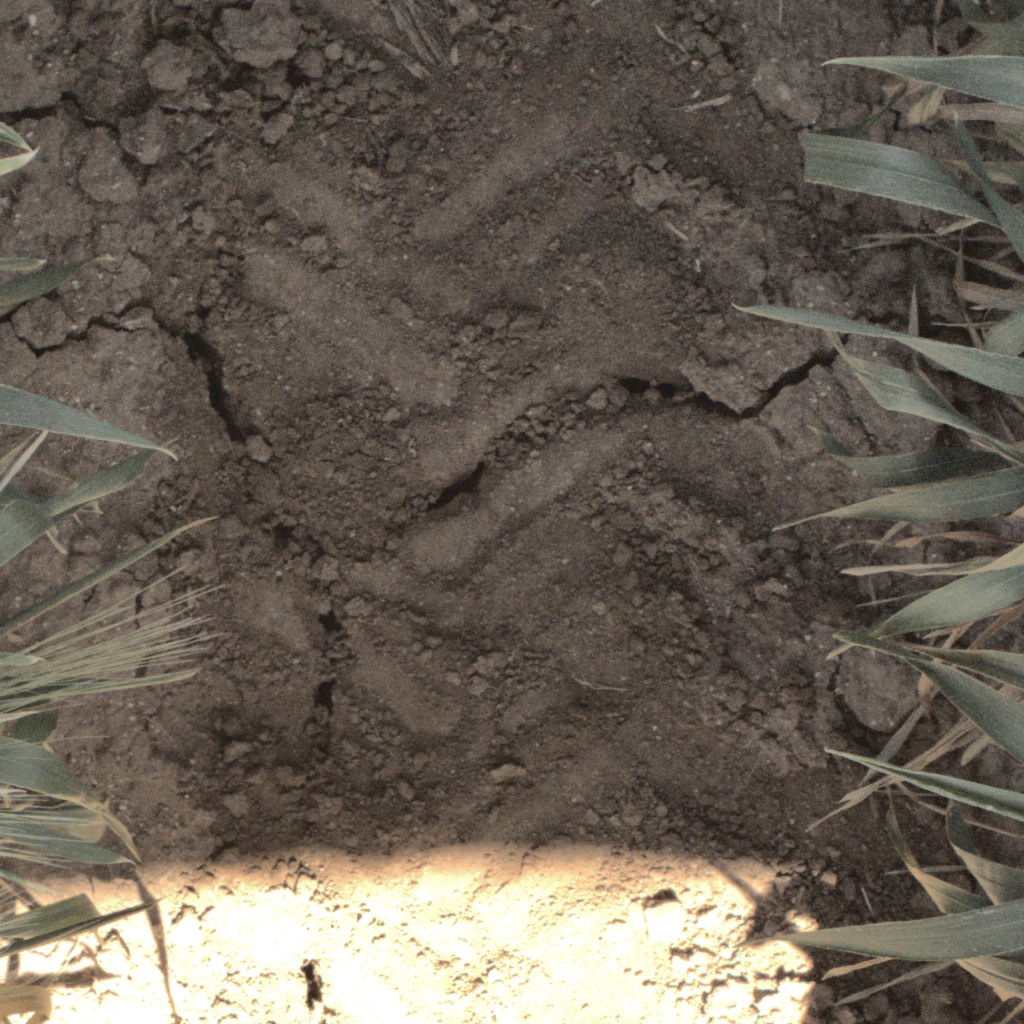}& \includegraphics[width=.25\linewidth]{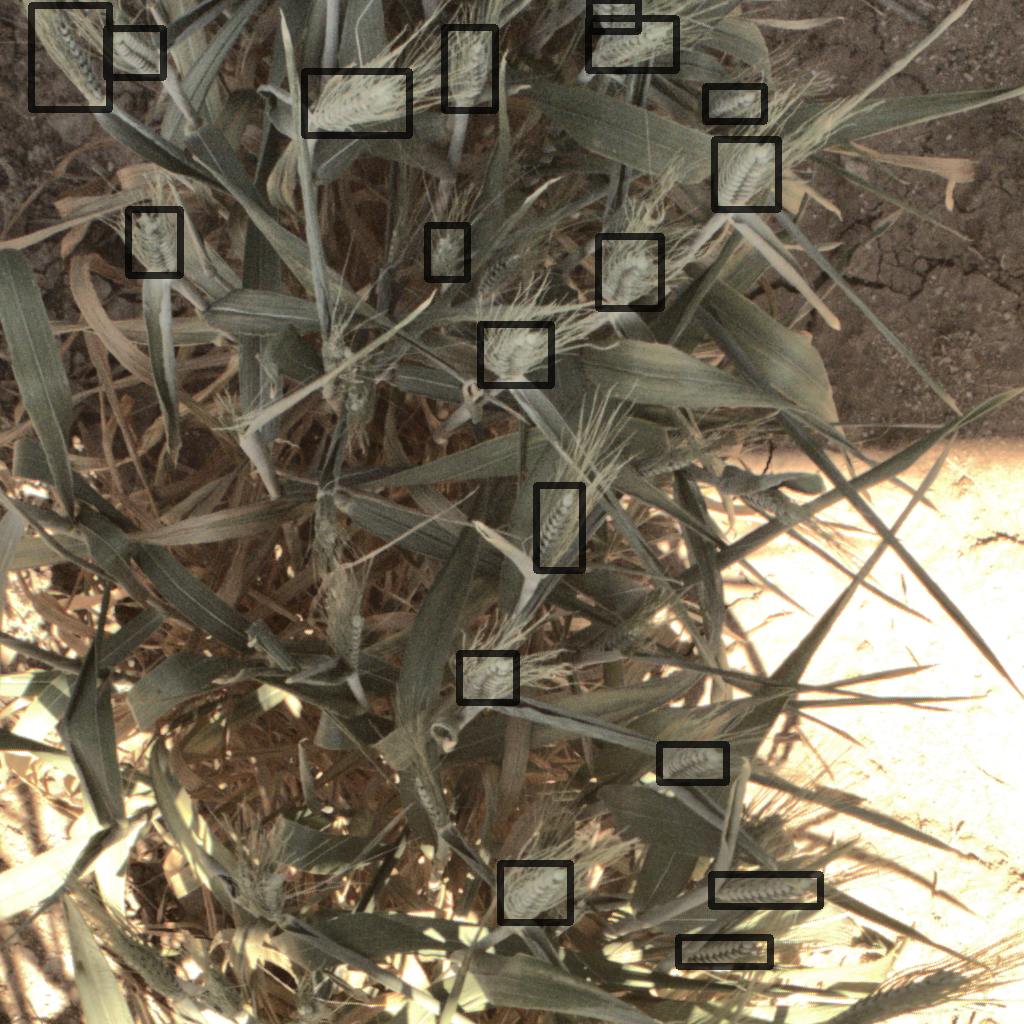}& \includegraphics[width=.25\linewidth]{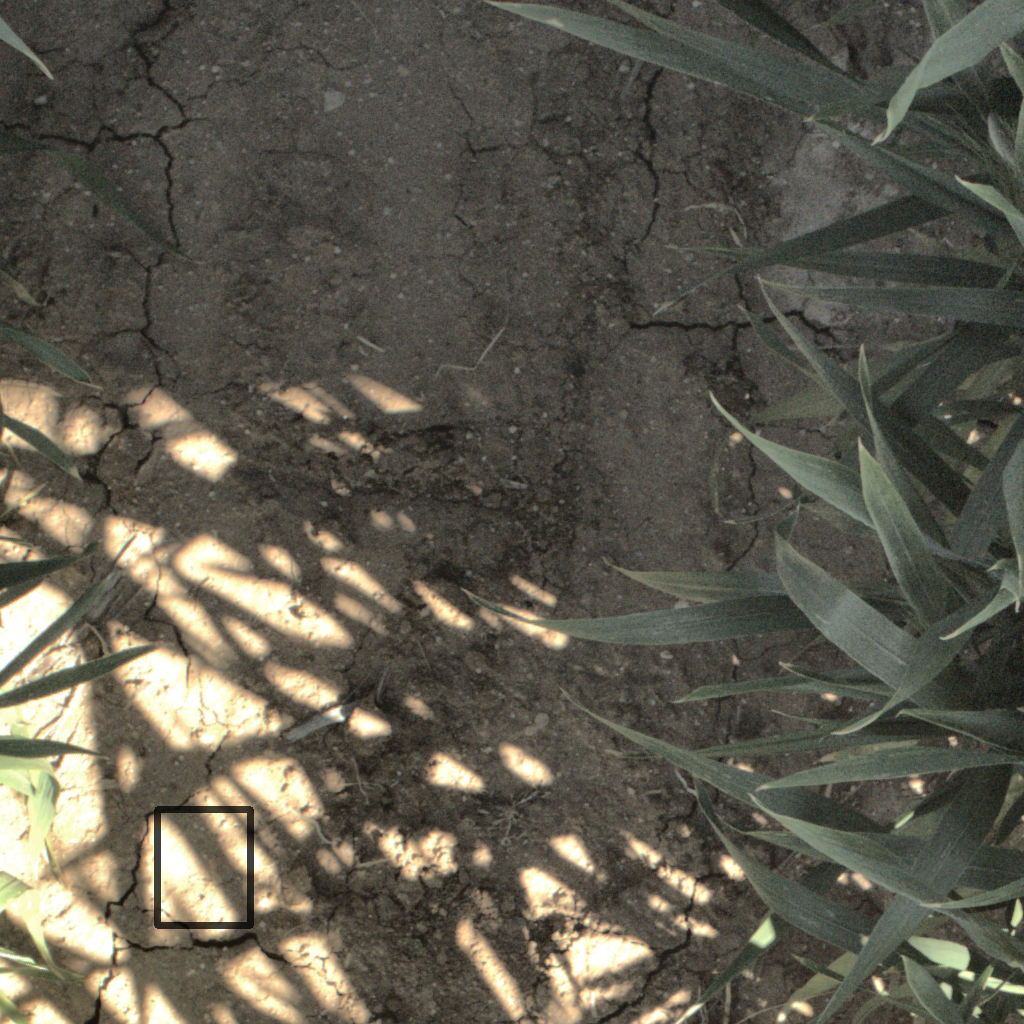}\\ \hline

                \textbf{Combined Ensemble}&\includegraphics[width=.25\linewidth]{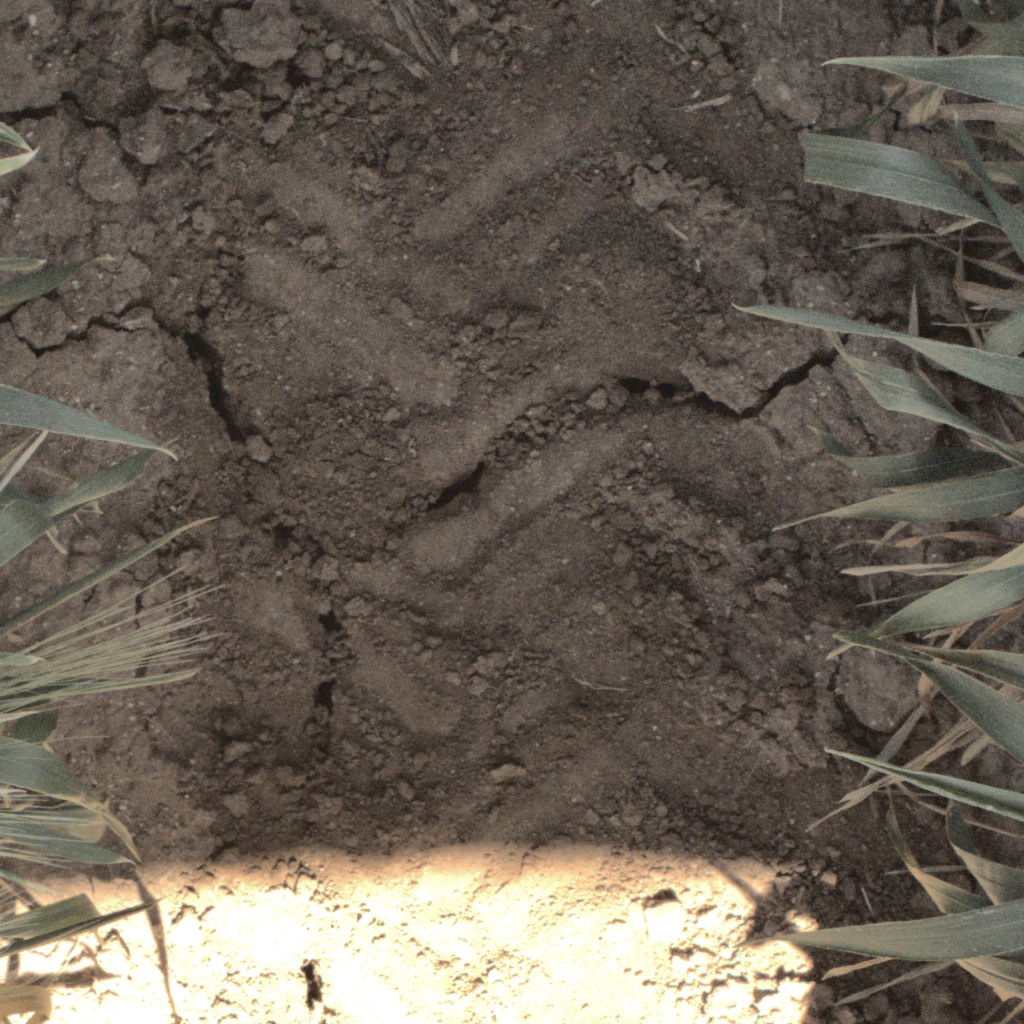}&\includegraphics[width=.25\linewidth]{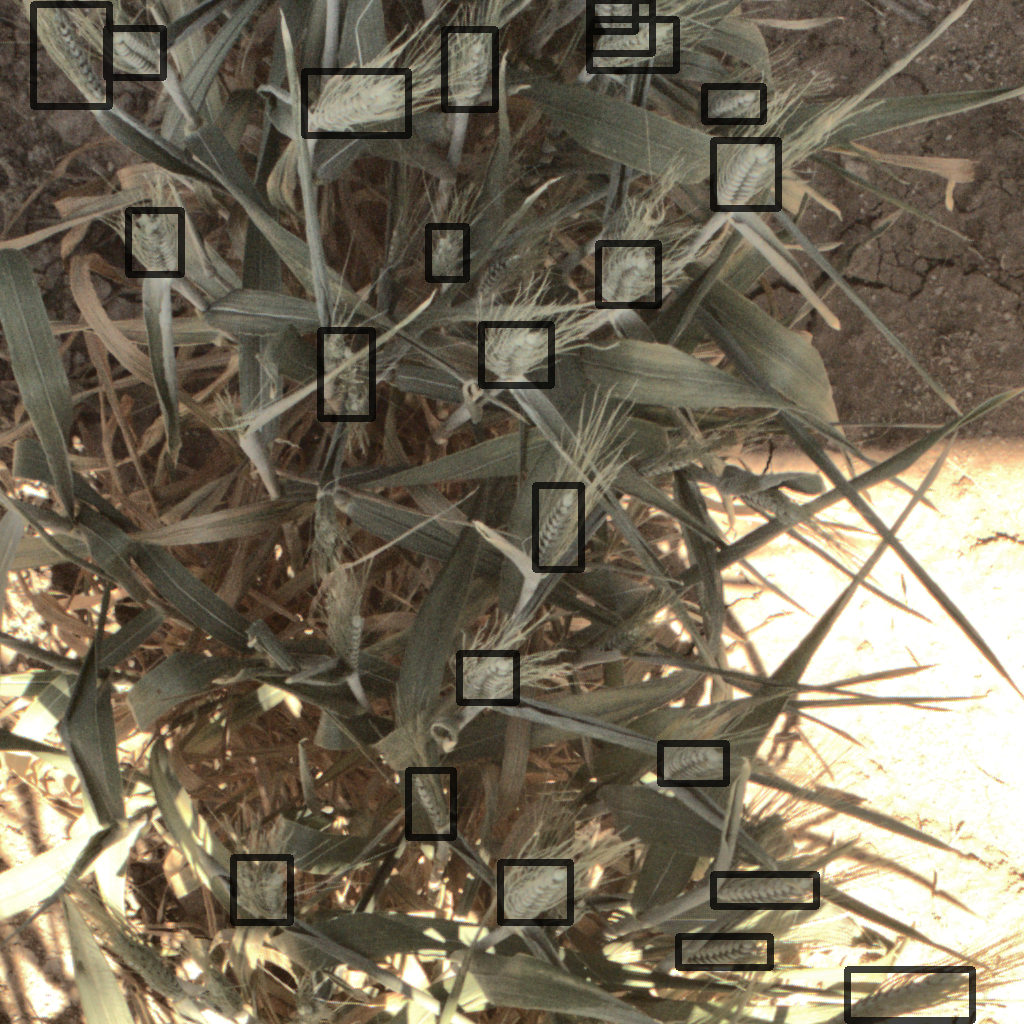}&\includegraphics[width=.25\linewidth]{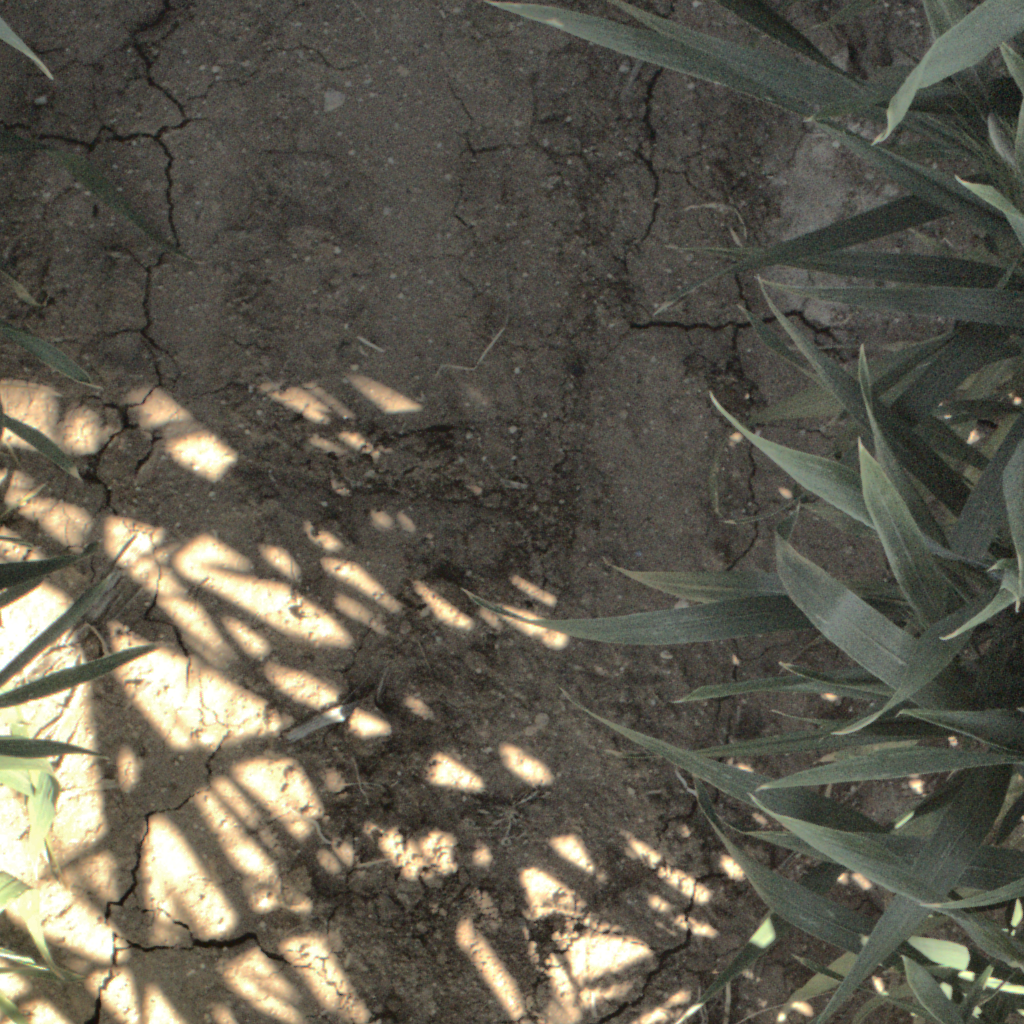}\\ \hline
            \end{tabular}
            \label{fig:final}
        \end{figure}

\subsection{Qualitative Results}
The qualitative comparison of various experiments and the predictions of the original YOLOv5 is shown in Fig.~\ref{fig:final}. As emphasised by the empirical results presented in table \ref{tab:overall}, weighted box fusion(WBF) with all sets of predictions combined shows the best performance. As we can see in the images, it has no miss-classifications due to high sunlight regions showing that the cause of miss-classification has been corrected without any further fine-tuning of the model.

\begin{figure}
    \centering
    \caption{Predictions indicating True Positives are not affected by the proposed approach.}
    \begin{tabular}{cc}
         \includegraphics[width=0.50\linewidth]{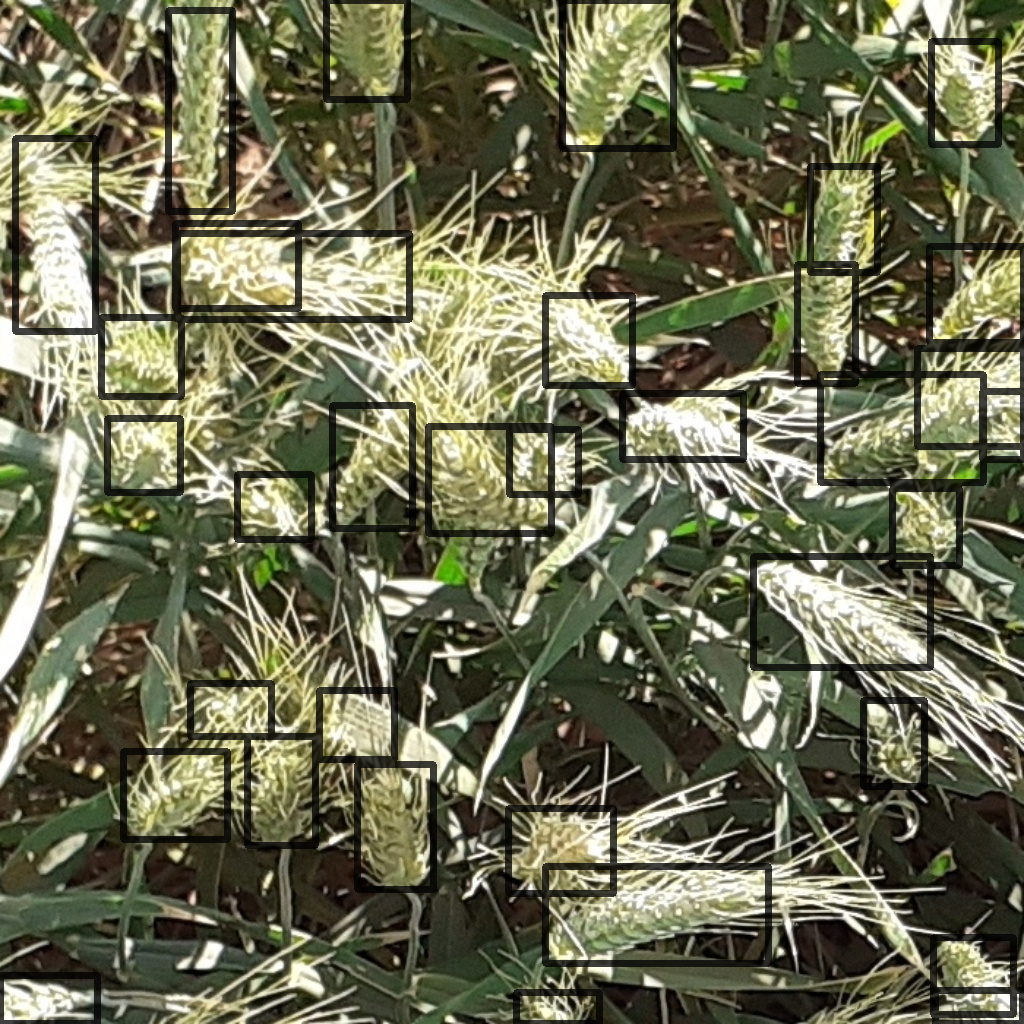}&
         \includegraphics[width=0.50\linewidth]{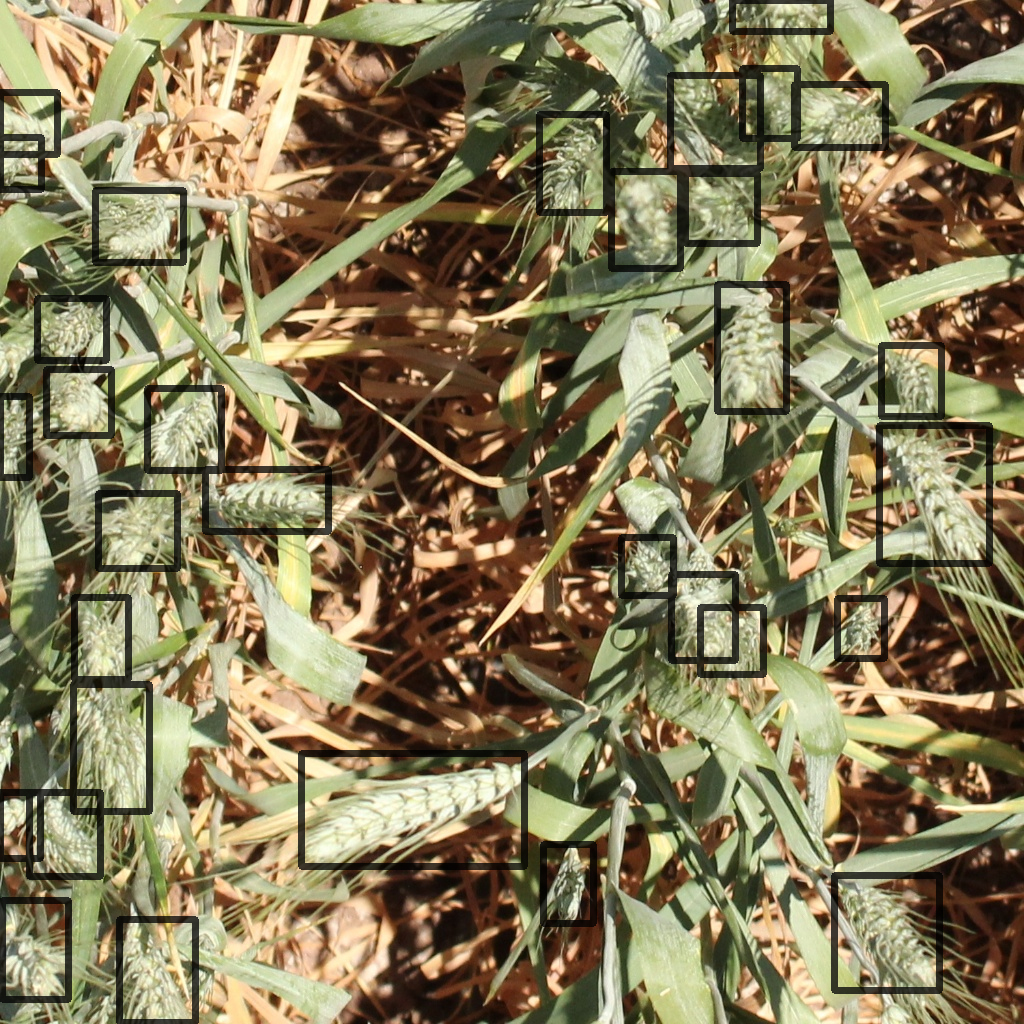}
    \end{tabular}
     \label{fig:false_pos}
\end{figure}

    In fig.~\ref{fig:final}, we can observe the gradual progression of improvement in performance with each experiment. The rows represents the experiments conducted on each sample while the columns represent the different samples. The number of mis-classifications due to spurious correlations are considerably high on the samples inferred using a plan YOLOv5 that was initially trained on the GWHD, 2021 train set. These miss-classifications reduce gradually and are almost eliminated after using weighted box fusion on the predictions. Further, we also show in fig.~\ref{fig:false_pos} that our proposed approach does not affect the true-positives predictions in bright images. We can see that despite the presence of sunlight, our framework detects all the wheat heads in the test set further validating the effectiveness of our proposed approach. 

\section{Conclusion}\label{sec:conclusion}
In this paper, we investigate and propose a solution to suppress the spurious correlations that exists in the Global Wheat Head Detection (GWHD) 2021 dataset. By training an autoencoder based architecture for YoloV5s backbone,  we observe that the trained model identifies the bright sunlight regions as false wheat head detections due to the spectral characteristics associated with the training wheat head instances. We propose a three step procedure to suppress the effects of these spurious correlations. The bright sunlight regions are initially identified by using thresholding operation followed by Navier-Stokes based inpainting. The artefacts from inpainting are suppressed by using a SegNet inspired autoencoder architecture followed by weighted boxes fusion operation to obtain the final set of detections. The proposed approach is observed to produce an overall improvement of 2$\%$ of ADA scores over the baseline and also effectively suppresses the spurious correlations present in GWHD 2021 dataset.

\bibliographystyle{unsrt} 
\FloatBarrier 
\bibliography{references}

\end{document}